\documentclass[]{article}
\usepackage[a4paper]{geometry}
\usepackage{mtsummit2023}
\usepackage{times}
\usepackage{url}
\usepackage{latexsym}
\usepackage{natbib}
\usepackage{layout}
\usepackage[hang,marginal]{footmisc}

\usepackage{afterpage}
\usepackage{subcaption}
\usepackage[dvipsnames]{xcolor}
\usepackage{xcolor}
\usepackage{anyfontsize}
\usepackage{microtype}
\usepackage{booktabs}
\usepackage{makecell}
\usepackage[normalem]{ulem}
\usepackage{caption}
\usepackage{booktabs}
\usepackage{algorithm}
\usepackage{algpseudocode}
\usepackage{appendix}
\usepackage{graphicx}
\usepackage{hyperref}
\usepackage{float}
\usepackage{amsmath}
\newcommand{\header}[1]{\vspace{1mm}\noindent\textbf{#1}.}
\usepackage{inconsolata}

\begin{document}

\title{\bf Joint Dropout: Improving Generalizability \\ in Low-Resource Neural Machine Translation through Phrase Pair Variables}

\author{\name{\bf Ali Araabi} \hfill  \addr{a.araabi@uva.nl}\\
       \name{\bf Vlad Niculae} \hfill \addr{v.niculae@uva.nl}\\
       \name{\bf Christof Monz} \hfill \addr{c.monz@uva.nl}\\
        \addr{Informatics Institute, University of Amsterdam,
        Amsterdam, The Netherlands}
}

\maketitle
\pagestyle{empty}

\begin{abstract}
Despite the tremendous success of Neural Machine Translation~(NMT), its performance on low-resource language pairs still remains subpar, partly due to the limited ability to handle previously unseen inputs, i.e., generalization. In this paper, 
we propose a method called \emph{Joint Dropout}, that addresses the challenge of low-resource neural machine translation by substituting phrases with variables, resulting in significant enhancement of compositionality, which is a key aspect of generalization.
We observe a substantial improvement in translation quality for language pairs with minimal resources, as seen in BLEU and Direct Assessment scores. Furthermore, we conduct an error analysis, and find Joint Dropout to also enhance generalizability of low-resource NMT in terms of robustness and adaptability across different domains.
\end{abstract}

\section{Introduction}
\label{section1}
Although Neural Machine Translation~(NMT) has made remarkable advances~\citep{VaswaniSPUJGKP17}, 
it still requires large amounts of data to induce correct generalizations that characterize human intelligence~\citep{lake_ullman_tenenbaum_gershman_2017}.
However, such a vast amount of data to make robust, reliable, and fair predictions is not available for low-resource NMT~\citep{KoehnK17}.

The generalizability of NMT has been extensively studied in prior research, revealing the volatile behaviour of translation outputs when even a single token in the source sentence is modified~\citep{BelinkovB18, FadaeeM20, LiYCZ20}. For instance, in the sentence ``\emph{smallpox killed billions of people on this planet}'' from our IWSLT test set, when replacing the noun ``\emph{smallpox}'' with another acute disease like ``\emph{tuberculosis}'', the model should ideally generate a correct translation by only modifying the relevant part while keeping the rest of the sentence unchanged. However, in many instances, such a small perturbation adversely affects the translation of the entire sentence, highlighting the limited generalization and robustness of existing NMT models \citep{FadaeeM20}.

Compositionality is regarded as the most prominent form of generalization that embodies the ability of human intelligence to generalize to new data, tasks, and domains~\citep{schmidhuber1990towards,LakeB18}, while other types mostly focus on the practical considerations across domains, tasks, and languages, model robustness, and structural generalization~\citep{abs-2210-03050}. 
Research in compositional generalization has two main aspects: evaluating the current models' compositional abilities as well as improving them. 

In terms of evaluation, some studies use artificially created test sets that mimic arithmetic-like compositionality~\citep{LakeB18}, while others evaluate compositionality in a more natural way~\citep{KeysersSSBFKMSS20,KimL20,DankersBH22}. 
In terms of improvement, earlier work aimed to enhance the models' compositional abilities on tasks such as semantic parsing datasets~\citep{QiuSPNLST22}, math word problem solving~\citep{abs-2209-01352}, data-to-text generation~\citep{MehtaRTKPS22}, and classification~\citep{KimRAO20}.
As for NMT, previous work has shown shortcomings in systematic compositional skills~\cite{LakeB18,LiYCZ20}, particularly for low-resource languages~\cite{DankersBH22}, yet no direct improvements have been proposed.

We aim to improve compositionality in NMT,
with a focus on low-resource scenarios that necessitate more robustness to form new combinations of previously seen smaller units.
To achieve this, we introduce Joint Dropout (JD), a simple and effective method that jointly replaces translation-equivalent phrase pairs in the source and target sentences with variables, encouraging the model to maintain the translation of the remaining sentence, regardless of the dropped phrases.
JD is orthogonal to and compatible with other methods for improving NMT performance. Specifically, it is designed to be data-centric and model-agnostic, allowing it to be easily combined with existing techniques that focus on different aspects of the NMT pipeline.

Our analysis on simulated and real low-resource data demonstrates JD's ability to significantly improve compositional generalization and translation quality.

\section{Methodology}
\label{section2}
Generalization has been a longstanding concern in the field of machine translation. In the past, Statistical MT utilized phrases as the fundamental translation units in order to consider contextual information, such as in Phrase-Based Statistical Machine Translation~\cite[PBSMT]{ZensON02}. To increase generalization, Hierarchical PBSMT proposed by \citet{Chiang05} builds upon the bilingual phrase pairs of PBSMT to learn hierarchical rules, capturing discontinuous translation equivalences and therefore allowing for better generalization.

Similarly, JD leverages bilingual phrases to make the rest of the translation not dependent on a specific phrase pair. However, the main idea behind JD originates from compositionality:
the meaning of a 
sentence is a function of the meanings of its
known atoms and how they are systematically and syntactically combined~\citep{partee1984compositionality}. 
By substituting \emph{meaning} with \emph{translation} in this definition, we come up with a rule of compositionality for translation systems:
\begin{equation}
    \tau (P \:\circ \: Q)=\tau (P)\; \circ \; \tau (Q)
\label{eq1}
\end{equation}
in which $\tau$ is the translation function, $P$ and $Q$ are the constituents of the sentence, and $\circ$ is a combiner. 
JD aims to transfer the principle of compositionality to the translation model in order to improve generalization and robustness of NMT by replacing  joint phrases with variables.
To exemplify, given the De-En sentence pair $\langle$\emph{Sie hat Rom besucht, She visited Rome}$\rangle$, we replace nouns with variables: $\langle$\emph{$X_1$ hat $X_2$ besucht, $Y_1$ visited $Y_2$}$\rangle$. 
Per Equation\ref{eq1}:
\begin{equation}
\small
    \begin{aligned}
        &\quad\;\tau(\textrm{Sie hat Rom besucht})\\
         &=\tau(((X_1 \:\textrm{hat}\: X_2 \:\textrm{besucht})\circ_{X_1}\:\textrm{Sie})\:\circ_{X_2}\textrm{Rom})\\
          &=\tau((X_1 \:\textrm{hat}\: X_2 \:\textrm{besucht})\circ_{X_1}\textrm{Sie})\circ_{\tau(X_2)}\tau(\textrm{Rom})\\
         &= (\tau(X_1 \:\textrm{hat}\: X_2 \:\textrm{besucht})\circ_{\tau(X_1)}\tau(\textrm{Sie}))\circ_{\tau(X_2)}\tau(\textrm{Rom})\\
         &=((Y_1 \: \textrm{visited} \: Y_2)\circ_{Y_1}\:\textrm{She})\circ_{Y_2}\textrm{Rome}\\
         &=\textrm{She visited Rome}
    \end{aligned}
\end{equation}
where $\tau(X_i)=Y_i$, and $\sigma \circ_{X_i} \gamma = \sigma[X_i \backslash \gamma]$, i.e., $\circ_{X_i}$ performs the replacement of $\gamma$ in the position $X_i$ in the sentence $\sigma$.
In the above sketch, we disregard any potential dependencies within the sentence. However, the variables are independent of the rest of the sentence in any manner.
Therefore, our goal is to enable the model to translate the entire sentence without being affected by the specific words or phrases at position $X_i$.
Hence, if the model learns the rules of composition properly, changing one or more lexical units will not hurt the rest of the translation.  
To this end, inspired by hierarchical PBSMT, 
we make use of bilingual phrases to improve generalization in low-resource NMT. However, since NMT has a strong capability to learn ordering through the cross-attention mechanism~\citep{ToralS17}, our aim is not to directly apply hierarchical PBSMT to NMT, but to propose an approximation as a lightweight and efficient regularization method. 

First, using Eflomal~\citep{OstlingT16},
an efficient word alignment tool, we generate symmetrized word alignments for the parallel training corpus to find the correspondences between source and target words in each pair of training sentences. Then, we use alignments as the input to generate the phrase translation table
by decomposing the source and target sentences into a set of dozens of bilingual phrase pairs that are consistent with the word alignment~\citep{KoehnOM03}.
In the next step, we select phrase pairs from the phrase table for each pair of training sentences and replace them with joint variables of~($X_i$,$Y_i$). More specifically, given a pair of sentences $S=\{w_1,w_2,...w_n\}$ and $T=\{w'_1,w'_2,...w'_m\}$, after substitution the sentences are $S=\{w_1,...,X_i,...,w_l,...,X_j,...,w_n\}$ and $T=\{w'_1,...,Y_i,...,w'_k,...,Y_j,...,w'_m\}$,
where $X$ and $Y$ are variables corresponding to the source and target phrases, respectively. 
We discuss different criteria to replace phrases with variables in \S\ref{section3.2}.\footnote{The code is available at \url{https://github.com/aliaraabi/Joint_Dropout}}
Finally, we add the variable-induced corpus to the original training set, effectively doubling its size.\footnote{We ensure all models undergo the same maximum number of updates during training, allowing a fair evaluation.}

\section{Experiments}
In this section, we present a comprehensive overview of our experiments. We begin by providing details regarding the datasets used and the training systems employed. Next, we delve into the specific criteria we considered when replacing phrases with variables. Subsequently, we discuss the significant improvements achieved by our proposed method, JD, across various aspects, including compositional generalization, translation performance, robustness, and the ability to generalize across domains.

\subsection{Experimental setup}
\header{Data} For the preliminary experiments, we use the TED data from the IWSLT 2014 German-English~(De-En) shared translation task~\citep{CettoloNSBF14} and randomly sample 
from the training data to represent various low-resource settings. In order to evaluate the models trained on IWSLT subsets, we use the concatenation of the IWLST $2014$ dev sets (tst$2010$–$2012$, dev$2010$, dev$2012$) as our test set, which consists of 6,750 sentence pairs.

We further evaluate JD on multiple actual low-resource language pairs: Belarusian~(Be), Galician~(Gl), and Slovak~(Sk) TED talks~\citep{QiSFPN18} and Slovenian~(Sl) from IWSLT2014~\citep{CettoloNSBF14} with training sets ranging from 4.5k to 55k sentence pairs. 

In order to evaluate the compositional ability of JD, following~\citet{DankersBH22}, we use an English-Dutch~(En-Nl) training set from OPUS~\footnote{Available on \url{https://github.com/Helsinki-NLP/Tatoeba-Challenge/blob/master/data/README-v2020-07-28.md}}~\citep{TiedemannT20} and randomly sample to create low-resource sets. To evaluate these models, we use both the ‘dev’ and the ‘devtest’ sets from  FLORES-101~\citep{GoyalGCCWJKRGF22} as the validation and test data.

\begin{table}[]
  \centering
  \begin{minipage}[t]{0.4\linewidth}
        \centering
  	\begin{tabular}{ll rr}
		\toprule 
		 \multicolumn{1}{l}{Setup} &
		 \multicolumn{1}{c}{\#Phrases} & 
		 \multicolumn{1}{c}{BLEU}  \\
		\midrule
		T-base &	$ 0 $ &	$ 12.2 $  \\
		T-opt. &	$0$ &	$18.0$\\
		\midrule
		 T-opt. + JD\_PP & $8013$ & $18.6$ \\
		 T-opt. + JD\_VP & $8013$ & $18.8$ \\
		 T-opt. + JD\_NP & $8013$ & $18.7$ \\
		 T-opt. + JD\_Mix & $8013$ & $18.8$ \\
		\bottomrule			
	\end{tabular}
    \captionof{table}{Results of Transformer-base, Transformer-optimized and Joint Dropout with various phrase types on 10K De-En training samples. Noun Phrases~(NP), Prepositional Phrases~(PP), Verb Phrases~(VP), and mixture (Mix) of all the above.}
  \label{tab6}
    \end{minipage}
    \hspace{0.1\linewidth}
    \begin{minipage}[t]{0.4\linewidth}
        \centering
  	\begin{tabular}{ll r}
		\toprule 
		Setup &
		BLEU  \\
		\midrule
		T-opt. &		$18.0$  \\
		T-opt. + JD &		$19.9$\\
		\midrule
		T-opt. + target variables only &  $15.5$ \\
		T-opt. + source variables only &  $17.3$ \\
		T-opt. + not aligned variables &  $17.8$ \\
		\bottomrule		
	\end{tabular}
	\caption{Importance of jointly dropping aligned phrases for model trained on 10K De-En samples.}
	\label{tab7}
    \end{minipage}
  
\end{table}
\header{Pre-processing} We apply punctuation normalization, tokenization, data cleaning, and true-casing using the Moses scripts~\citep{KoehnHBCFBCSMZDBCH07}. The sentence length is limited to a maximum of 175 tokens during training. After replacing phrases with variables, we also apply BPE segmentation~\citep{SennrichHB16a} with the parameter tailored to the low-resource training data~\citep{AraabiM20}. We ensure that variables are not split into smaller segments.

\header{Data annotation} To generate a realistic test set for evaluating robustness against sentence perturbation, we first randomly select 300 translation outputs from the inference stage of baseline systems trained using optimized parameters on 20k samples. These outputs are then ranked using the Direct Assessment~(DA) method by engaging native annotators. The top 100 outputs are then selected and the corresponding outputs from the model trained with JD are extracted and ranked using DA. Next, the input sentences are modified by replacing specific phrases or words while ensuring their syntactic and semantic accuracy. After obtaining the outputs for both the baseline and JD systems on the perturbed sentences, we conduct a DA on them.

\header{Training system} To conduct our experiments, we employ two different models: Transformer-optimized~\citep{AraabiM20}, specifically tailored to low-resource NMT and Transformer-base with its default hyper-parameters~\citep{VaswaniSPUJGKP17}. This choice allows us to demonstrate that the improvements achieved are consistent and independent of the specific model settings.
We use the Fairseq library~\citep{OttEBFGNGA19} for our experiments and average sacreBLEU\footnote{sacreBLEU parameters:
nrefs:1$|$case:mixed$|$eff:no$|$tok:13a$|$smooth:exp$|$version:2.0.0}~\citep{Post18} over three runs as the evaluation metric. All of the models are trained on a single GPU for a few hours with the model parameters ranging from 28M to 47M.

\subsection{Joint Dropout parameters}
\label{section3.2}
The following conditions are considered in replacing phrases with variables. First, we do not allow two adjacent phrases to be replaced with variables.
Although phrases can vary in length, we consider all phrases as potential candidates for substitution with variables, irrespective of their length. 
After conducting initial experiments, we have determined that setting the maximum number of variables allowed in each sentence to $10$ yields satisfactory results.

Since noun phrases are the most cross-linguistically common phrases, we hypothesize that they are good candidates to be replaced.
Therefore, in a set of experiments we investigate the choice of phrase types. We consider four different scenarios: replacing 1)~only Noun Phrases~(NP), 2)~only Prepositional Phrases~(PP), 3)~only Verb Phrases~(VP), and 4)~mixtures of all the above.
We train four systems on $10$k samples from the TED talks dataset with four different substitution scenarios yet the same number of variables~(8013).\footnote{8013 is the number of all possible substitutions for PPs.}
We use the constituency parser from Stanford CoreNLP~\citep{manning-EtAl:2014:P14-5}.

It is important to note that our selection of phrase pairs in both languages is solely based on English constituency parse trees. We do not rely on the use of a constituency parser, which is often not available for many low-resource languages. The results presented in Table~\ref{tab6} demonstrate that the choice of different phrase types does not lead to significant differences in our method. Therefore, our approach eliminates the need for a constituency parser, making it applicable to a wider range of low-resource languages.
\begin{figure}[t]
\centering
  \includegraphics[width=10cm]{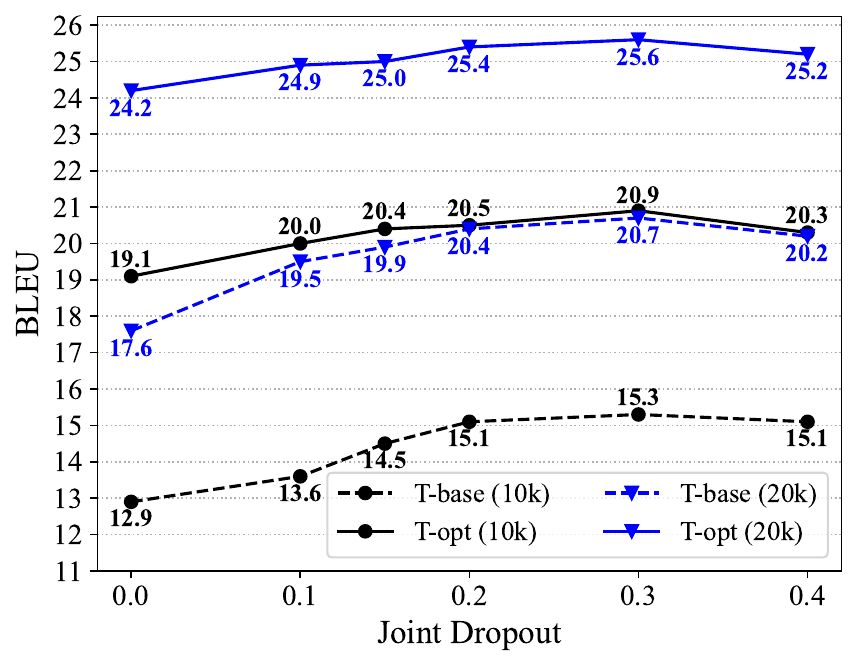}
  \caption{Effect of different Joint Dropout rates on Transformer-base and Transformer-optimized, on the validation sets of two De-En training subsets.}
  \label{fig1}
\end{figure}
For the rest of the experiments, we substitute phrases regardless of their types.

To make JD independent of a phrase translation table, we consider not-aligned phrases in both or either translation sides. The importance of using aligned phrases is demonstrated in Table~\ref{tab7}, where it is observed that utilizing not-aligned phrases results in a degradation of performance by $2.5$ BLEU points. This finding highlights the significance of incorporating aligned phrases in the JD method.

To maintain control over the number of variables across the entire training corpus, we introduce a concept called the \emph{Joint Dropout rate}. This rate is determined by calculating the proportion of dropped tokens, specifically from within phrases, in relation to the total length of both the source and target sentences. By utilizing this Joint Dropout rate, we can effectively regulate and manage the presence of variables throughout the training process.
Figure~\ref{fig1} illustrates the improvements achieved by two distinct systems as the Joint Dropout rate increases. Notably, JD consistently improves the performance of both the Transformer-base and Transformer optimized models. Specifically, on a dataset of 110k samples, JD yields a notable increase of $+2.4$ BLEU points for the Transformer-base model and $+1.8$ BLEU points for the Transformer-optimized model. Moreover, when evaluating a larger dataset of 20k samples, JD further improves translation quality by $+3.1$ BLEU points for the Transformer-base model and $+1.4$ BLEU points for the Transformer-optimized model.

We see that
the Joint Dropout rate of $0.3$ 
is a good choice%
, while more noise in the training set hurts performance.
We use this rate for the remainder of the experiments. 

\begin{table}[t]
	\centering
	\begin{tabular}{r rrrr}
		\toprule
            & \multicolumn{2}{c}{BLEU} & \multicolumn{2}{c}{Consistency} \\
            \cmidrule(lr){2-3}
            \cmidrule(lr){4-5}
		\multicolumn{1}{l}{\#Samples} &
		\multicolumn{1}{l}{T-opt.} &
            \multicolumn{1}{l}{T.opt.+JD} &
		\multicolumn{1}{c}{T-opt.}   & \multicolumn{1}{c}{T.opt.+JD }\\
		\midrule
		$5k$ & $4.2$ & $\mathbf{6.1}$ &$2.0$& $\mathbf{4.0^\ast}$\\
		  $20k$ & $10.4$ & $\mathbf{10.7}$ &$8.1$ & $\mathbf{11.0^\ast}$\\
            $40k$ & $12.8$ & $\mathbf{13.4}$ &$13.1$ & $\mathbf{15.6^\ast}$\\
		$80k$& $\mathbf{16.4}$ & $\mathbf{16.4}$ &$37.1$ & $\mathbf{43.8^\ast}$\\
		  $200k$ & $\mathbf{19.2}$ & $18.7$ & $58.2$ & $\mathbf{65.4^\ast}$\\
		\bottomrule			
	\end{tabular}
	\caption{BLEU and consistency scores (En~$\rightarrow$ Nl) when replacing a noun 
 in the subject position with a different noun.
 Significant improvements on compositionality of JD over the strong baseline are marked with $\ast$ (approximate randomization, p $<$ 0.01).}
	\label{tab9}
\end{table}

\subsection{Compositional generalization}
Unlike phenomena
such as idioms,
which require a more global understanding,
JD concentrates on improving compositionality at the local level. In this section, we aim to evaluate our method on local compositionality. Here, we take advantage of the most relevant theoretically grounded test from~\citet{HupkesDMB20} which is \emph{systematicity}, a notion frequently used in the context of compositionality. This attribute of the model concerns the recombination of known parts and rules, ensuring that the model's ability to grasp novel inputs is systematically tied to their aptitude to comprehend related inputs. For instance, understanding ``smallpox killed billions of people on this planet'' and ``tuberculosis'', also implies understanding ``tuberculosis killed billions of people on this planet''. 

Given that there are an infinite number of potential novel combinations that can be derived from known parts in natural data, we concentrate on a sentence-level, context-free rule: S~$\rightarrow$ NP VP, as proposed by~\citet{DankersBH22}, where a noun from the NP in the subject position is replaced with a different noun, while maintaining number agreement with the VP. 
Additionally, they highlight that a systematic system necessitates consistency. We assess this systematicity of translations based on their consistency across various contexts when presenting words or phrases. Consistency is measured by evaluating the equality between two translations while taking into account anticipated modifications. In S~$\rightarrow$~NP~VP setup, after replacement, translations are deemed consistent if there is only one word difference between them. Table~\ref{tab9} illustrates that JD consistently enhances the consistency scores for various low-resource data conditions.

\subsection{Translation performance}
In this section, we conduct a comprehensive evaluation of translation quality across multiple language pairs to assess the effectiveness of JD. The results presented in Table~\ref{tab1} highlight the significant improvements in translation quality achieved by JD for actual low-resource language pairs.
Importantly, these improvements also hold true for the reverse language direction.

Furthermore, we compare JD to three comparable methods for dropping tokens: Zero-Out, where the embedding of a token is set to zero~\citep{Sennrich16}, Token Drop, which replaces tokens with the \texttt{<dropped>} tag~\cite{ZhangQDZ20}, and SwitchOut, where words are replaced with random words from their corresponding vocabularies~\cite{WangPDN18}.
The results in Table~\ref{tab2-a} demonstrate that Zero-Out only provides marginal improvements. Moreover, both Token Drop and SwitchOut methods prove to be ineffective in low-resource scenarios. In contrast, JD consistently outperforms these methods, particularly in extreme low-resource cases. 
As shown in Table~\ref{tab2-a}, Zero-Out only provides marginal improvements. In addition, while Token Drop and SwitchOut methods prove to be ineffective in low-resource situations, JD consistently yields the largest improvements, especially for extreme low-resource cases. In addition, Table~\ref{tab2-b} provides additional evidence supporting the superiority of JD over similar methods, even when optimized parameters for the Transformer model are not specifically chosen.

\begin{table}[]
	\centering
	\begin{tabular}{l rrrr | rrrr}
		\toprule 
		 \multicolumn{1}{l}{Method} &
		 \multicolumn{1}{c}{Be-En} & 
		 \multicolumn{1}{c}{Gl-En} & 
		 \multicolumn{1}{c}{Sl-En} &
		 \multicolumn{1}{c}{Sk-En} &
          \multicolumn{1}{c}{En-Be} & 
		 \multicolumn{1}{c}{En-Gl} & 
		 \multicolumn{1}{c}{En-Sl} &
		 \multicolumn{1}{c}{En-Sk}\\
		\midrule
		T-base &	$ 4.6 $ &	$13.4$  & $8.9$   & $24.0$ & $ 3.5 $ &	$10.1$  & $6.8$   & $19.0$ \\
		T-base + JD & $\mathbf{6.5}$ &   $\mathbf{15.8}$  & $\mathbf{10.2}$  & $\mathbf{25.0}$  & $\mathbf{4.5}$ &   $\mathbf{12.9}$  & $\mathbf{7.8}$  & $\mathbf{19.2}$ \\
		\midrule
		T-opt. &	$8.0$   &	$21.8$  & $15.2$  & $28.9$ & $5.5$   &	$18.3$  & $12.3$  & $23.1$ \\
		T-opt. + JD & $\mathbf{9.9}$ &   $\mathbf{22.8}$  & $\mathbf{16.1}$  & $\mathbf{29.8}$  & $\mathbf{7.3}$ &   $\mathbf{18.9}$  & $\mathbf{12.7}$  & $\mathbf{23.5}$\\

		\bottomrule			
	\end{tabular}
	\caption{BLEU scores for actual extremely low-resource languages: Be, Gl, Sl, and Sk with 4.5k, 10k, 13k, and 55k training samples, respectively. %
}
	\label{tab1}
\end{table}
\begin{table}[t]
\small
	\centering
        \begin{minipage}[b]{0.4\linewidth}
            \centering
            \begin{tabular}{l rrr}
            
		\toprule 
		\multicolumn{1}{l}{Method} &
		\multicolumn{1}{c}{5k} & 
		\multicolumn{1}{c}{10k} & 
		\multicolumn{1}{c}{20k} \\
		\midrule
		T-opt.     &$13.4$ &$18.0$   &	$23.0$ \\
		T-opt. + ZO & $13.6$ &$ 18.3 $ &	$22.8$  \\
		T-opt. + TD & $9.5$	&$ 16.8 $ &	$23.9$  \\
		T-opt. + SW & $13.4$	&$ 18.4 $ &	$24.0$  \\
		T-opt. + JD & $\mathbf{15.2}$ &$\mathbf{19.9}$ &   $\mathbf{24.4}$\\
		\bottomrule			
	\end{tabular}
        \subcaption{Transformer-optimized}
	\label{tab2-a}    
    \end{minipage}
    \hspace{0.1\linewidth}
    \begin{minipage}[b]{0.4\linewidth}
            \centering
            \begin{tabular}{l rrr}
            
		\toprule 
		\multicolumn{1}{l}{Method} &
		\multicolumn{1}{c}{5k} & 
		\multicolumn{1}{c}{10k} & 
		\multicolumn{1}{c}{20k} \\
		\midrule
		T-base      & $8.6$	&$ 12.1 $ &	$16.6$  \\
		T-base + ZO & $8.9$	&$ 13.3 $ &	$18.3$  \\
		T-base + TD & $5.3$	&$ 8.9 $ &	$14.6$  \\
		T-base + SW & $5.5$	& $9.8$ &	$14.5$  \\
		T-base + JD & $\mathbf{9.8}$ &$\mathbf{14.5}$ &   $\mathbf{19.1}$  \\
		\bottomrule			
	\end{tabular}
        \subcaption{Transformer-base}
	\label{tab2-b} 
    \end{minipage}
\caption{Comparing BLEU scores for Joint Dropout~(JD) and the reimplementations of Token Drop~(TD), Zero Out~(ZO), and SwitchOut~(SW) on 5k, 10k and 20k training samples from IWSLT De-En.}	
\label{tab2} 
\end{table}

\subsection{NMT Robustness}
Recent work has shown that trivial modifications to the source sentence can cause unexpected changes in the translation~\citep{FadaeeM20}. Furthermore, models with stronger compositional abilities are anticipated to generate more robust translations~\cite{DankersBH22}. To evaluate the robustness of JD against such modifications, we differ from previous methods that automatically introduce noise to the test set~\citep{MichelN18, ChengJM19} which is prone to creating semantic and syntactic errors in the input. Instead, we manually develop a more realistic test set. 

\begin{table}[t]
\small
	\centering
	\begin{tabular}{ll rrr}
		\toprule 
		Method &
		Metric &
		Orig.  & 
		Per. &
		$\Delta$\\
		\midrule
		T-base  & \makecell[l]{\small{DA}\\\small{BLEU}} &	\makecell[r]{$62.1$ \\ $28.5$ } &	\makecell[r]{$49.3$\\$26.0$} & \makecell[r]{$-12.8$\\$-2.5$}    \\
		
		\cmidrule{2-5}
		
		T-base + JD & \makecell[l]{\small{DA}\\\small{BLEU}} &\makecell[r]{$69.8$\\$30.7$} &	\makecell[r]{$59.3$\\$30.4$}  & \makecell[r]{$\mathbf{-10.5}$\\$\mathbf{-0.3}$}    \\
		
		\midrule
		
		T-opt. &   \makecell[l]{\small{DA}\\\small{BLEU}}    &	\makecell[r]{$79.9$\\$37.4$}   &	\makecell[r]{$56.6$\\$31.8$}  & \makecell[r]{$-23.3$\\$-5.6$}   \\
		
		\cmidrule{2-5}
		
		T-opt. + JD & \makecell[l]{\small{DA}\\\small{BLEU}} & \makecell[r]{$83.7$\\$41.8$}   &   \makecell[r]{$77.4$\\$39.9$}  & \makecell[r]{$\mathbf{-6.3}$\\$\mathbf{-1.9}$} \\
		
		\bottomrule			
	\end{tabular}
	
	\caption{Direct assessment and BLEU scores, pre and post input perturbation on random samples from De-En test set.}
	
	\label{tab4}
\end{table}

\begin{table*}[t]
	\centering
	\begin{tabular}{l@{~}l@{~}l}
		\toprule 
		 \multicolumn{1}{l}{} &
		 \multicolumn{1}{l}{\small{Original test sentence}} & 
		 \multicolumn{1}{l}{\small{Test sentence after perturbation}}  \\
		\midrule
		\small{Src} &	\makecell[l]{\small{[\textbf{ein Kind in Indien}] sagt:}\\ \small{``heute habe ich einen Affen gesehen''.}} & 
        \makecell[l]{\small{\{\textbf{meine Oma in China}\} sagt:} \\ \small{``heute habe ich einen Affen gesehen''.}} \\[2.5ex]
	\small{Ref.} & \makecell[l]{\small{[a child in India] says} ,\\ \small{``\uline{I saw a monkey} today .''}}	  &	
        \makecell[l]{\small{\{my grandmother in China\} says,} \\
        \small{``\uline{I saw a monkey} today .''}}   \\
		\midrule
		
		\small{T-opt.} &	\makecell[l]{\small{[a child in India] says,} \\ \small{``today \uline{I've seen a monkey}s.''}} &	\makecell[l]{\small{\{my grandmother's mother in China\}}\\ \small{ says, \small{``\uwave{\textcolor{red}{Look}} today.''}}}\\[2.5ex]
		
		\small{T-opt. + JD} & \makecell[l]{\small{[a kid in India] says,} \\  \small{``\uline{I've seen a monkey}s today.''}} &
        \makecell[l]{\small{\{my grandmother in China\} says,} \\ \small{ ``today \uline{\textcolor{OliveGreen}{I've seen a monkey}}s.''}} \\
		\bottomrule			
	\end{tabular}
	\vspace{-2pt}
	\caption{By replacing the German noun phrase \textbf{\emph{ein Kind in Indien}} [a child in India] with \textbf{\emph{meine Oma in China}} [my grandmother in China], there is no undesirable behavior in the rest of the translation when using Joint Dropout. Underlined text means the rest of the translation is approximately the same with the reference, while the wavy underline means it has changed. Bracket shows the phrase that we perturb, while the curly bracket is the perturbed phrase}
	\label{tab3}
\end{table*}
First, based on Direct Assessment~(DA) on a $100$-point scale~\citep{GrahamBMZ13}, we select the top $100$ sentences out of randomly selected $300$ translation outputs generated by a Transformer-optimized model trained on 20k samples. We then alter the input sentences by replacing a specific phrase or word, while ensuring that they remain syntactically and semantically accurate. Table~\ref{tab4} illustrates that perturbing the original sentences results in a smaller performance decrease for the model trained with JD, when compared to the baseline.
This means that our proposed method significantly decreases the volatile behavior of low-resource NMT.

Table~\ref{tab3} shows an example of perturbing a sentence. After replacing ``\emph{ein Kind in Indien}'' in the source sentence with ``\emph{meine Oma in China}'',  while the rest of the translation is negatively affected using the baseline model, the JD shows more robustness against the input perturbation and does not exhibit any negative behavior.

\subsection{Generalization across domains}
In low-resource language settings, NMT systems frequently encounter challenges when it comes to achieving effective translation across distinct domains. This is primarily attributed to their tendency to prioritize the idiosyncrasies of the training domain, rather than capturing the broader linguistic characteristics shared by the language pairs.
Therefore, in addition to evaluating generalization in terms of compositionality and robustness, it is also crucial to assess generalization concerning distributional shift and uncertainty estimation~\citep{abs-2210-03050}. While the definition of a domain is not precisely defined~\citep{WeesBWM15},for our evaluation, we consider TED talks and news as belonging to different domains.

Table~\ref{tab5} provides insights into the behavior of JD when there is a domain shift between the training domain (TED talks) and the test domain (news from WMT). The results demonstrate that JD exhibits greater robustness in such scenarios, showcasing its ability to better handle distributional shifts and improve translation quality across different domains. This highlights the effectiveness of JD in mitigating the negative effects of domain-specific training and enhancing the generalizability of NMT systems in low-resource language pairs.

\begin{table}[t]
	\centering
	\begin{tabular}{l rrr}
		\toprule
		\multicolumn{1}{l}{Method} & \multicolumn{1}{c}{10k}   & \multicolumn{1}{c}{20k} & \multicolumn{1}{c}{40k}\\
		\midrule
		T-base          & $2.4$& $3.9$  & $7.1$   \\
		T-base + JD &$\mathbf{3.2}$ & $\mathbf{5.4}$  & $\mathbf{9.8}$   \\
		\midrule
		T-opt.          &$6.2$ & $8.7$  & $13.9$  \\
		T-opt. + JD & $\mathbf{7.5}$ & $\mathbf{10.9}$ & $\mathbf{14.6}$ \\
		\bottomrule			
	\end{tabular}
	\caption{Results of training on different subsamples of TED talks and testing on a domain with different distribution~(Newstest2020).}
	\label{tab5}
\end{table}
\section{Conclusion}
Despite the fact that NMT's success is closely tied to having large amounts of training data, it is still beneficial to explore methods that can help improve generalization when working with limited data. In this paper, we introduce Joint Dropout as a straightforward yet effective approach to enhancing the compositional generalization and translation quality  of low-resource NMT. Specifically, we demonstrate that jointly replacing phrases with variables has a regularizing effect that mitigates overfitting by enabling the system to translate sentences regardless of the specific phrases present at the variable positions.

\section{Future work}
We only focus on improving generalizability of low-resource NMT, while higher-resource settings might also gain from joint variables. Additionally, we demonstrate the effectiveness of our proposed method using multiple low-resource language pairs, whereas there are many other language pairs with limited data. Furthermore, since JD tries to capture the rules of compositionality in translation, we expect more benefit to the language pairs with less similarity. Additionally, our approach is data-centric and model-agnostic, applicable to various models and tasks beyond the methods evaluated in this paper. Therefore, it has the potential to improve existing pre-trained models such as mBART~\citep{LiuGGLEGLZ20}, when fine-tuning on low-resource languages, but further experimentation is needed to confirm its effectiveness.
We leave these investigations to future work.

\section{Broader Impact}
The implementation of NMT has brought about significant progress in the translation field, however, it also poses potential challenges such as liability for mistakes made by using NMT and mistranslation, which could be more of a concern when dealing with limited data. Furthermore, the high ability of NMT to generalize well presents a potential risk of difficulty in identifying errors, specifically those related to compositionality. This can be a concern in safety-critical domains where a single error can have severe consequences. Moreover, the ability of NMT to produce more coherent and fluent translations may impede the identification of where the system is malfunctioning, thus hindering the correction of errors or biases in the model.  

\bibliographystyle{apalike}
\bibliography{main}

\end{document}